\documentclass[11pt,a4paper]{article}
\usepackage{emnlp2018}
\usepackage{times}
\usepackage{latexsym}

\usepackage{microtype}
\usepackage{graphicx}
\usepackage{booktabs} 

\usepackage{amsmath}
\usepackage{amsfonts}
\usepackage{multirow}
\usepackage{url}
\usepackage{microtype}
\usepackage{subcaption}
\usepackage{marginnote}
\usepackage{bbm}
\usepackage{float}

\usepackage[shortlabels]{enumitem}
\usepackage{tikz}
\usepackage{ifthen}
\usepackage{stmaryrd}

\usepackage{wrapfig}
\usepackage{caption}

\usetikzlibrary{positioning}

\DeclareMathOperator*{\argmax}{argmax}

\newcommand{\st}[1]{\textbf{#1}}
\newcommand{\E}{\mathcal{E}}
\newcommand{\R}{\mathcal{R}}

\newcommand{\phiijkmod}{\psi_{i, k, j}^{t_{1} + t_{2} \rightarrow t}}
\newcommand{\phiijt}[3]{\psi_{#1, #2}^{#3}}

\aclfinalcopy 

\title{Neural Compositional Denotational Semantics for Question Answering}

\author{
    Nitish Gupta\thanks{\hspace{0.3em} Work done while interning with Facebook AI Research.} \\
    University of Pennsylvania \\
    Philadelphia, PA \\
    \texttt{nitishg@seas.upenn.edu}
    \And
    Mike Lewis \\
    Facebook AI Research \\
    Seattle, WA \\
    \texttt{mikelewis@fb.com}
}

\date{}

\begin{document}
\maketitle
\begin{abstract}
Answering compositional questions requiring multi-step reasoning is challenging.
We introduce an end-to-end differentiable model for interpreting questions about a knowledge graph (KG), which is inspired by formal approaches to semantics.
Each span of text is represented by a denotation in a KG and a vector that captures ungrounded aspects of meaning. Learned composition modules recursively combine constituent spans, culminating in a grounding for the complete sentence which answers the question. For example, to interpret ``\emph{not green}'', the model represents ``\emph{green}'' as a set of KG entities and ``\emph{not}'' as a trainable ungrounded vector---and then uses this vector to parameterize a composition function that performs a complement operation. For each sentence, we build a parse chart subsuming all possible parses, allowing the model to jointly learn both the composition operators and output structure by gradient descent from end-task supervision.
The model learns a variety of challenging semantic operators, such as quantifiers, disjunctions and composed relations, and infers latent syntactic structure. 
It also generalizes well to longer questions than seen in its training data, in contrast to RNN, its tree-based variants, and semantic parsing baselines.
\end{abstract}

\begin{figure*}[tb]
\setlength\abovecaptionskip{0pt}
\setlength\belowcaptionskip{-15pt}

    \centering
    
    \scalebox{0.7}{
    \includegraphics[width=\textwidth]{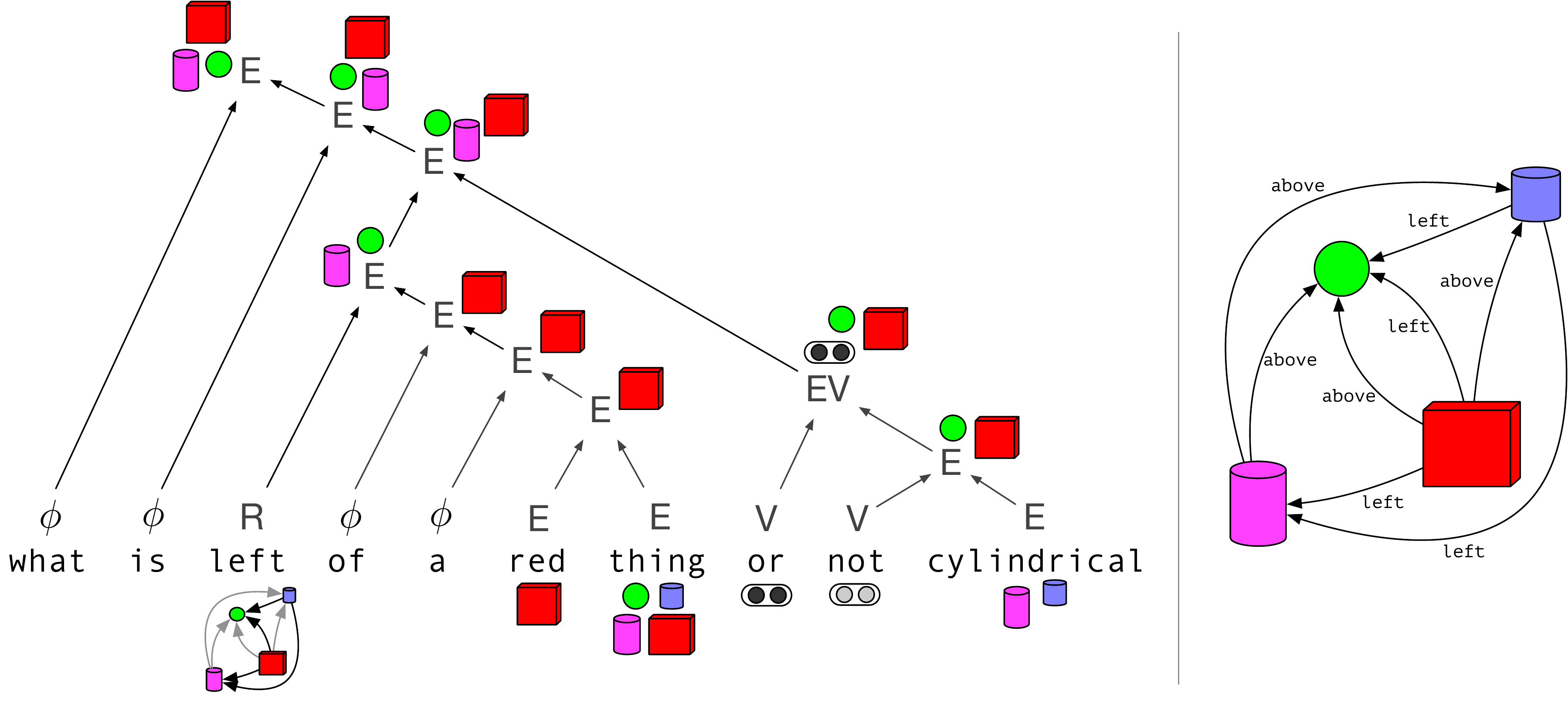}
    }
    \caption{{ A correct parse for a question given the knowledge graph on the right, using our model. We show the type for each node, and its denotation in terms of the knowledge graph. The words \emph{or} and \emph{not} are represented by vectors, which parameterize composition modules. The denotation for the complete question represents the answer to the question. 
    Nodes here have types $E$ for sets of entities, $R$ for relations, $V$ for ungrounded vectors, $EV$ for a combination of entities and a vector, and $\phi$ for semantically vacuous nodes.
    While we show only one parse tree here, our model builds a parse chart subsuming all trees.} }
    \label{fig:overview}
\end{figure*}

\section{Introduction}
\label{sec:intro}

Compositionality is a mechanism by which the meanings of complex expressions are systematically determined from the meanings of their parts, and has been widely assumed in the study of both artificial and natural languages \cite{Montague1973} as a means for allowing speakers to generalize to understanding an infinite number of sentences. Popular neural network approaches to question answering use a restricted form of compositionality, typically encoding a sentence word-by-word, and then executing the complete sentence encoding against a knowledge source \citep{perez2017}. Such models can fail to generalize from training data in surprising ways. Inspired by linguistic theories of compositional semantics, we instead build a latent tree of interpretable expressions over a sentence, recursively combining constituents using a small set of neural modules. Our model outperforms RNN encoders, particularly when test questions are longer than training questions.

Our approach resembles Montague semantics, in which a tree of interpretable expressions is built over the sentence, with nodes combined by a small set of composition functions. 
However, both the structure of the sentence and the composition functions are learned by end-to-end gradient descent. To achieve this, we define the parametric form of small set of composition modules, and then build a parse chart over each sentence subsuming all possible trees. Each node in the chart represents a span of text with a distribution over groundings (in terms of booleans and knowledge base nodes and edges), as well as a vector representing aspects of the meaning that have not yet been grounded. The representation for a node is built by taking a weighted sum over different ways of building the node (similar to \newcite{Maillard2017}). The trees induced by our model are linguistically plausible, in contrast to prior work on structure learning from semantic objectives \citep{williams:2018}.

Typical neural approaches to grounded question answering first encode a question with a recurrent neural network (RNN), and then evaluate the encoding against an encoding of the knowledge source (for example, a knowledge graph or image) \citep{SantoroRBMPBL17}.
In contrast to classical approaches to compositionality, constituents of complex expressions are not given explicit interpretations in isolation. 
For example, in \emph{Which cubes are large or green?}, an RNN encoder will not explicitly build an interpretation for the phrase \emph{large or green}.
We show that such approaches can generalize poorly when tested on more complex sentences than they were trained on.
Our approach instead imposes independence assumptions that give a linguistically motivated inductive bias. In particular, it enforces that phrases are interpreted independently of surrounding words, allowing the model to generalize naturally to interpreting phrases in different contexts. In our model, \emph{large or green} will be represented as a particular set of entities in a knowledge graph, and be intersected with the set of entities represented by the \emph{cubes} node.

Another perspective on our work is as a method for learning layouts of Neural Module Networks (NMNs) \citep{andreas2016neural}.
Work on NMNs has focused on construction of the structure of the network, variously using rules, parsers and reinforcement learning \citep{Andreas2016LearningTC, Hu2017LearningTR}. Our end-to-end differentiable model jointly learns structures and modules by gradient descent.

Our model is a new combination of classical and neural methods, which maintains the interpretability and generalization behaviour of semantic parsing, while being end-to-end differentiable.




\section{Model Overview}
\label{sec:model}
Our task is to answer a question $q=w_{1..|q|}$, with respect to a Knowledge Graph (KG) consisting of nodes $\E$ (representing entities) and labelled directed edges $\R$ (representing relationship between entities). In our task, answers are either booleans, or specific subsets of nodes from the KG.

Our model builds a parse for the sentence, in which phrases are grounded in the KG, and a small set of composition modules are used to combine phrases, resulting in a grounding for the complete question sentence that answers it. For example, in Figure~\ref{fig:overview}, the phrases \emph{not} and \emph{cylindrical} are interpreted as a function word and an entity set, respectively, and then \emph{not cylindrical} is interpreted by computing the complement of the entity set.
The node at the root of the parse tree is the answer to the question.
Our model answers questions by:
\paragraph{(a)} Grounding individual tokens in a KG, that can either be grounded as particular sets of entities and relations in the KG, as ungrounded vectors, or marked as being semantically vacuous. For each word, we learn parameters that are used to compute a distribution over semantic types and corresponding denotations in a KG (\S~\ref{ssec:semtype}).

\paragraph{(b)} Combining representations for adjacent phrases into representations for larger phrases, using trainable neural composition modules (\S~\ref{ssec:comprules}). This produces a denotation for the phrase.

\paragraph{(c)} Assigning a binary-tree structure to the question sentence, which determines how words are grounded, and which phrases are combined using which modules. We build a parse chart subsuming all possible structures, and train a parsing model to increase the likelihood of structures leading to the correct answer to questions. Different parses leading to a denotation for a phrase of type $t$ are merged into an expected denotation, allowing dynamic programming (\S~\ref{ssec:parsing}).

\paragraph{(d)} Answering the question, with the most likely grounding of the phrase spanning the sentence. 

\section{Compositional Semantics}

\subsection{Semantic Types}
\label{ssec:semtype}
Our model classifies spans of text into different semantic types to represent their meaning as explicit denotations, or ungrounded vectors.
All phrases are assigned a distribution over semantic types.
The semantic type determines how a phrase is grounded, and which composition modules can be used to combine it with other phrases.
A phrase spanning $w_{i..j}$ has a denotation $\llbracket w_{i..j} \rrbracket_{KG}^{t}$ for each semantic type $t$.
For example, in Figure \ref{fig:overview}, \emph{red} corresponds to a set of entities, \emph{left} corresponds to a set of relations, and \emph{not} is treated as an ungrounded vector. 

The semantic types we define can be classified into three broad categories.

\paragraph{Grounded Semantic Types:} Spans of text that can be fully grounded in the KG. 

\begin{enumerate}
    \item
    {\bf Entity} (\st{E}): Spans of text that can be grounded to a set of entities in the KG, for example, \emph{red sphere} or \emph{large cube}.
    \st{E}-type span grounding is represented as an attention value for each entity, $[p_{e_{1}}, \ldots, p_{e_{|\E|}}]$, where $p_{e_{i}} \in [0, 1]$.
    This can be viewed as a soft version of a logical set-valued denotation, which we refer to as a soft entity set.
    
    \item
    {\bf Relation} (\st{R}): Spans of text that can be grounded to set of relations in the KG, for example, \emph{left of} or \emph{not right of or above}.
    \st{R}-type span grounding is represented by a soft adjacency matrix $A \in \mathbb{R}^{|\E|\times|\E|}$ where $A_{ij} = 1$ denotes a directed edge from $e_{i} \rightarrow e_{j}$. 
    
    \item
    {\bf Truth} (\st{T}): Spans of text that can be grounded with a Boolean denotation, for example, \emph{Is anything red?}, \emph{Is one ball green and are no cubes red?}.
    \st{T}-type span grounding is represented using a real-value $p_{true} \in [0,1]$ that denotes the probability of the span being true.
\end{enumerate}

\paragraph{Ungrounded Semantic Types:} Spans of text whose meaning cannot be grounded in the KG.

\begin{enumerate}
\item
{\bf Vector} (\st{V}):
    This type is used for spans representing functions that cannot yet be grounded in the KG (e.g. words such as \emph{and} or \emph{every}).
    These spans are represented using $4$ different real-valued vectors $v_{1}$-$v_{4}$ $\in$ $\mathbb{R}^{2}$-$\mathbb{R}^{5}$, that are used to parameterize the composition modules described in \S \ref{ssec:comprules}. 
    
    \item
    {\bf Vacuous} ($\pmb{\phi}$): Spans that are considered semantically vacuous, but are necessary syntactically, e.g. \emph{of} in \emph{left of a cube}. During composition, these nodes act as identity functions. 
\end{enumerate}

\paragraph{Partially-Grounded Semantic Types:} 
Spans of text that can only be partially grounded in the knowledge graph, such as \emph{and red} or \emph{are four spheres}. Here, we represent the span by a combination of a grounding and vectors, representing grounded and ungrounded aspects of meaning respectively.
The grounded component of the representation will typically combine with another fully grounded representation, and the ungrounded vectors will parameterize the composition module.
We define 3 semantic types of this kind: \st{EV}, \st{RV} and \st{TV}, corresponding to the combination of entities, relations and boolean groundings respectively with an ungrounded vector. Here, the word represented by the vectors can be viewed as a binary function, one of whose arguments has been supplied.

\subsection{Composition Modules}
\label{ssec:comprules}
\begin{figure*}[ht!]
  \captionsetup[subfigure]{labelformat=empty}
  \begin{subfigure}[t]{.48\textwidth}
    \centering
    \includegraphics[width=0.6\linewidth]{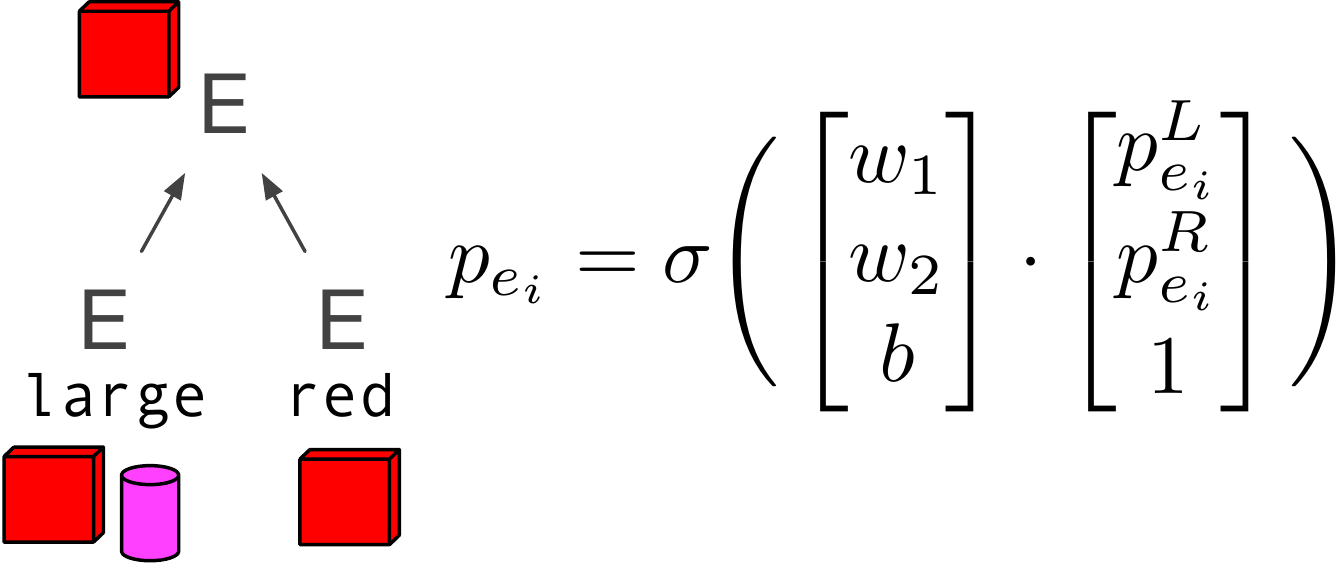}
    \caption{\label{fig:eee}{\bf \st{E} + \st{E} $\rightarrow$ \st{E}}: This module performs a function on a pair of soft entity sets, parameterized by the model's global parameter vector $[w_{1}, w_{2}, b]$ to produce a new soft entity set. The composition function for a single entity's resulting attention value is shown. Such a composition module can be used to interpret compound nouns and entity appositions. For example, the composition module shown above learns to output the intersection of two entity sets.}
  \end{subfigure}
  \hfill
  \begin{subfigure}[t]{.48\textwidth}
    \centering
    \includegraphics[width=0.6\linewidth]{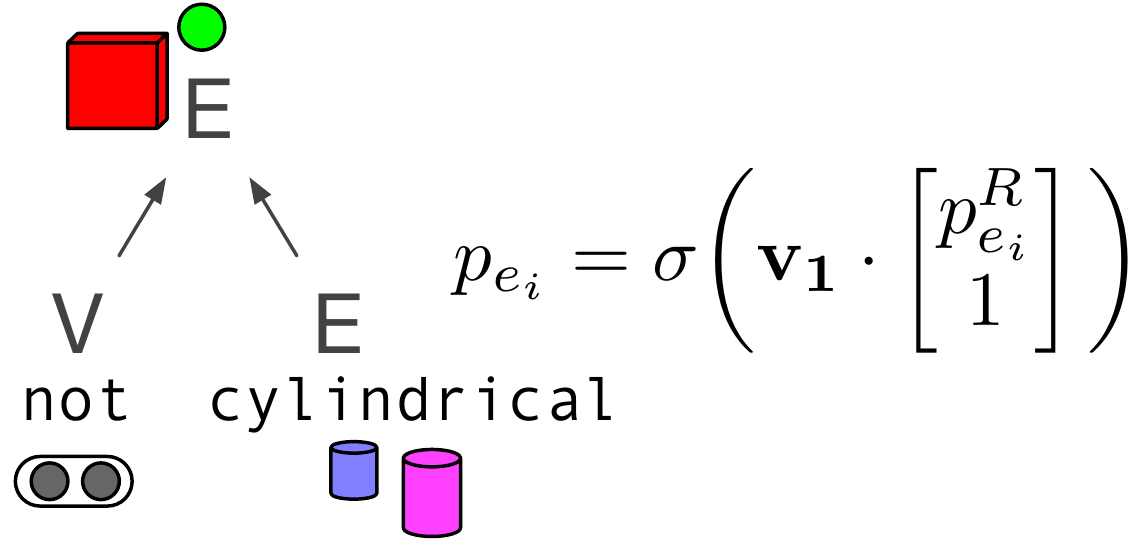}
    \caption{\label{fig:wee}{\bf \st{V} + \st{E} $\rightarrow$ \st{E}}: This module performs a function on a soft entity set, parameterized by a word vector, to produce a new soft entity set. For example, the word \emph{not} learns to take the complement of a set of entities. The entity attention representation of the resulting span is computed by using the indicated function that takes the $v_{1} \in \mathbb{R}^{2}$ vector of the \st{V} constituent as a          parameter argument and the entity attention vector of the \st{E} constituent as a function argument.
        }
  \end{subfigure}

    \medskip

  \begin{subfigure}[t]{.48\textwidth}
    \centering
    \includegraphics[width=0.7\linewidth]{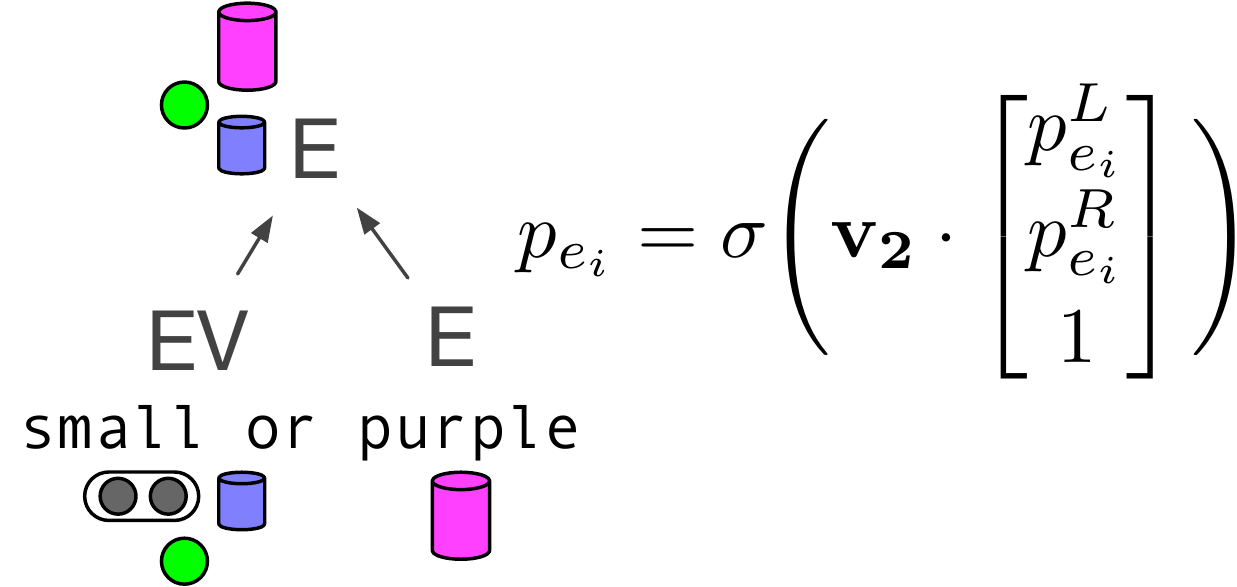}
    \caption{\label{fig:ewee}{\bf \st{EV} + \st{E} $\rightarrow$ \st{E}}: This module combines two soft entity sets into a third set, parameterized by the $v_{2}$ word vector. This composition function is similar to a linear threshold unit and is capable of modeling various mathematical operations such as logical conjunctions, disjunctions, differences etc. for different values of $v_{2}$. For example, the word \emph{or} learns to model set union. 
    }
  \end{subfigure}
  \hfill
  \begin{subfigure}[t]{.48\textwidth}
    \centering
    \includegraphics[width=0.7\linewidth]{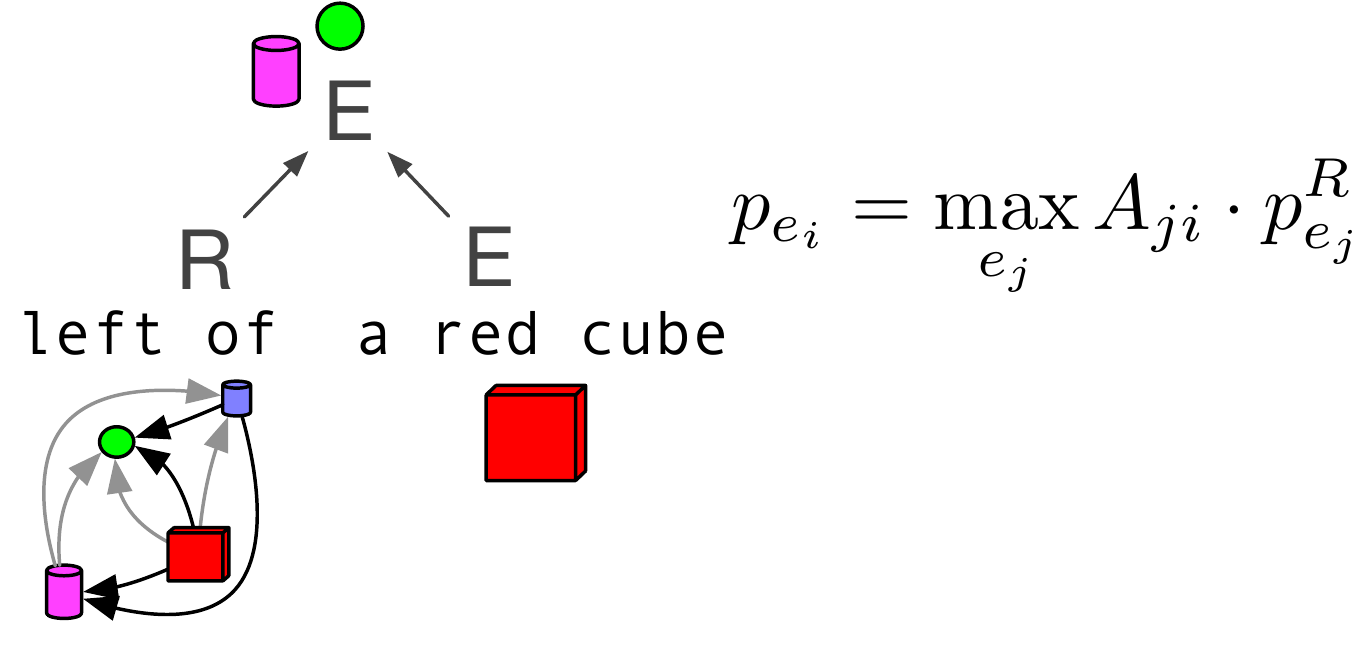}
    \caption{\label{fig:ree}{\bf \st{R} + \st{E} $\rightarrow$ \st{E}}: This module composes a set of relations (represented as a single soft adjacency matrix) and a soft entity set to produce an output soft entity set. The composition function uses the adjacency matrix representation of the \st{R}-span and the soft entity set representation of the \st{E}-span. }
  \end{subfigure}
  
    \medskip

  \begin{subfigure}[t]{.48\textwidth}
    \centering
    \includegraphics[width=0.95\linewidth]{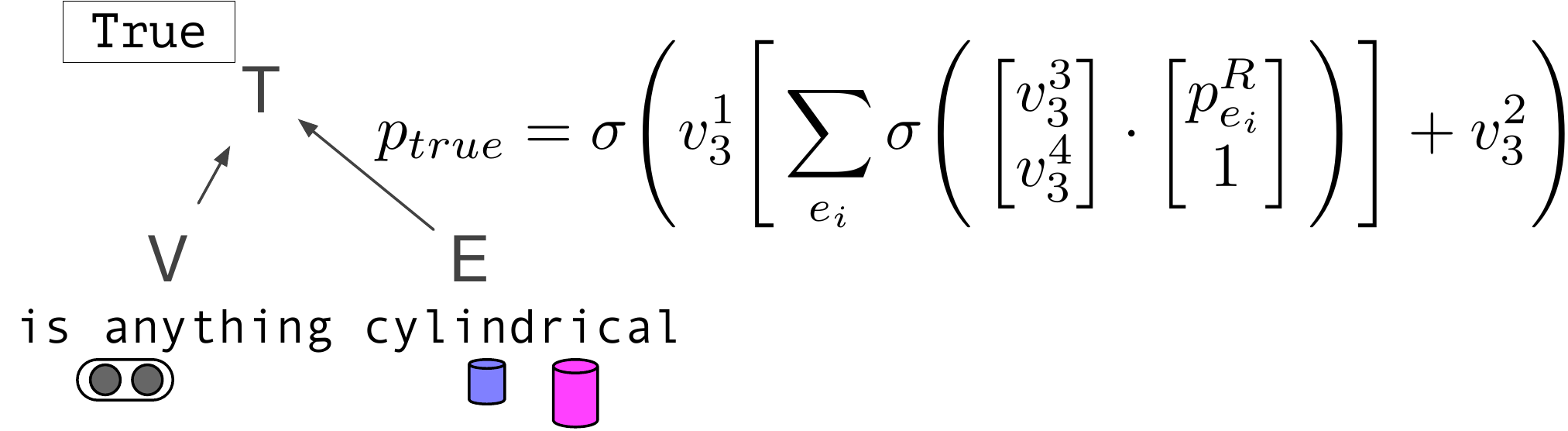}
    \caption{\label{fig:web}{\bf \st{V} + \st{E} $\rightarrow$ \st{T}}: This module maps a soft entity set onto a soft boolean, parameterized by word vector ($v_{3}$). 
    The module counts whether a sufficient number of elements are in (or out) of the set.
    For example, the word \emph{any} should test if a set is non-empty.
    }
  \end{subfigure}
  \hfill
  \begin{subfigure}[t]{.48\textwidth}
    \centering
    \includegraphics[width=0.95\linewidth]{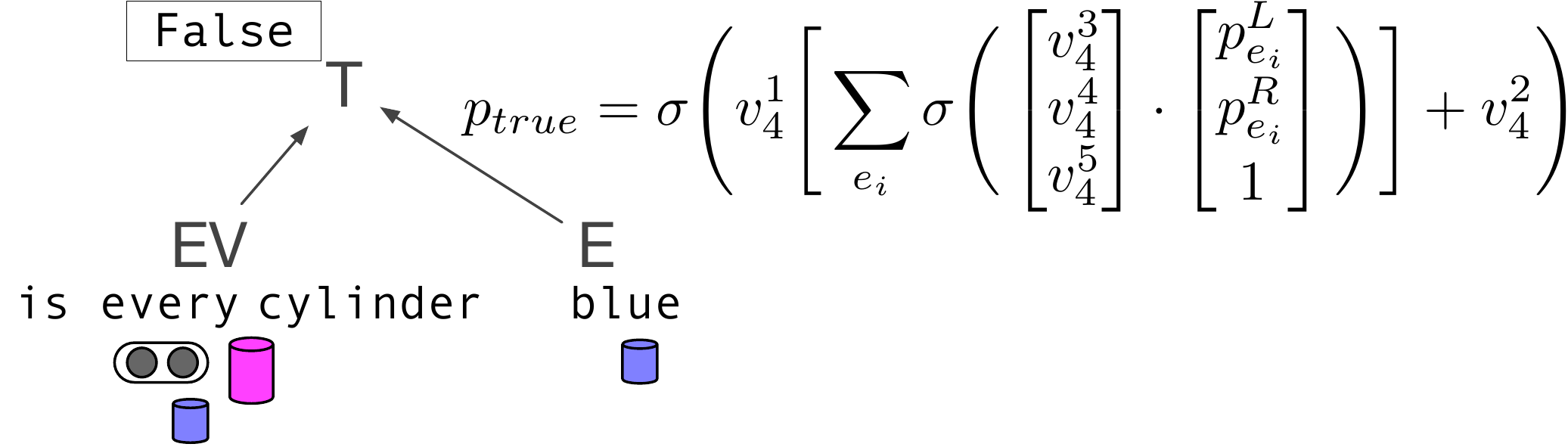}
    \caption{\label{fig:eweb}{\bf \st{EV} + \st{E} $\rightarrow$ \st{T}}: This module combines two soft entity sets into a soft boolean, which is useful for     modelling generalized quantifiers. For example, in \emph{is every cylinder blue}, the module can use the inner sigmoid to test if an element $e_i$ is in the set of cylinders ($p^{L}_{e_{i}}\approx 1$) but not in the set of blue things ($p^{R}_{e_{i}}\approx 0$), and then use the outer sigmoid to return a value close to 1 if the sum of elements matching this property is close to 0.  }
  \end{subfigure}

   \medskip

  \begin{subfigure}[t]{.48\textwidth}
    \centering
    \includegraphics[width=0.75\linewidth]{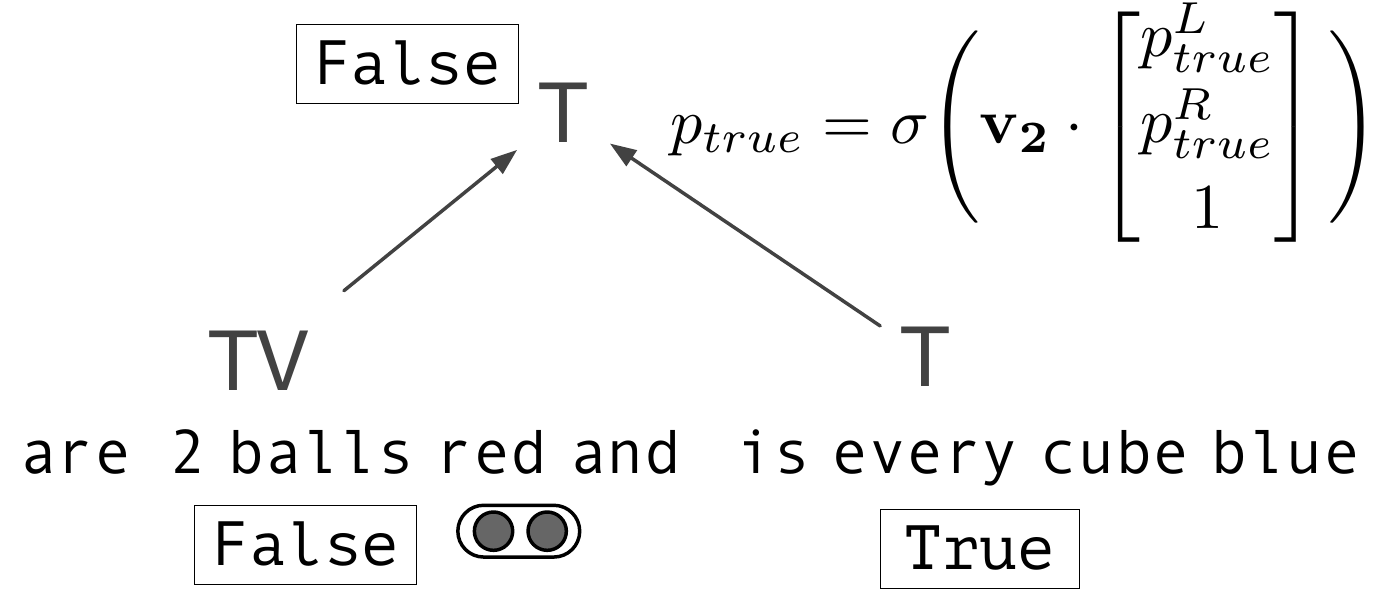}
    \caption{\label{fig:bwbb}{\bf \st{TV} + \st{T} $\rightarrow$ \st{T}}: This module maps a pair of soft booleans into a soft boolean using the $v_{2}$ word vector to parameterize the composition function. Similar to \st{EV}~+~\st{E}~$\rightarrow$~\st{E}, this module facilitates modeling a range of boolean set operations. Using the same functional form for different composition functions allows our model to use the same ungrounded word vector ($v_{2}$) for compositions that are semantically analogous.}
  \end{subfigure}
  \hfill
  \begin{subfigure}[t]{.48\textwidth}
    \centering
    \includegraphics[width=0.7\linewidth]{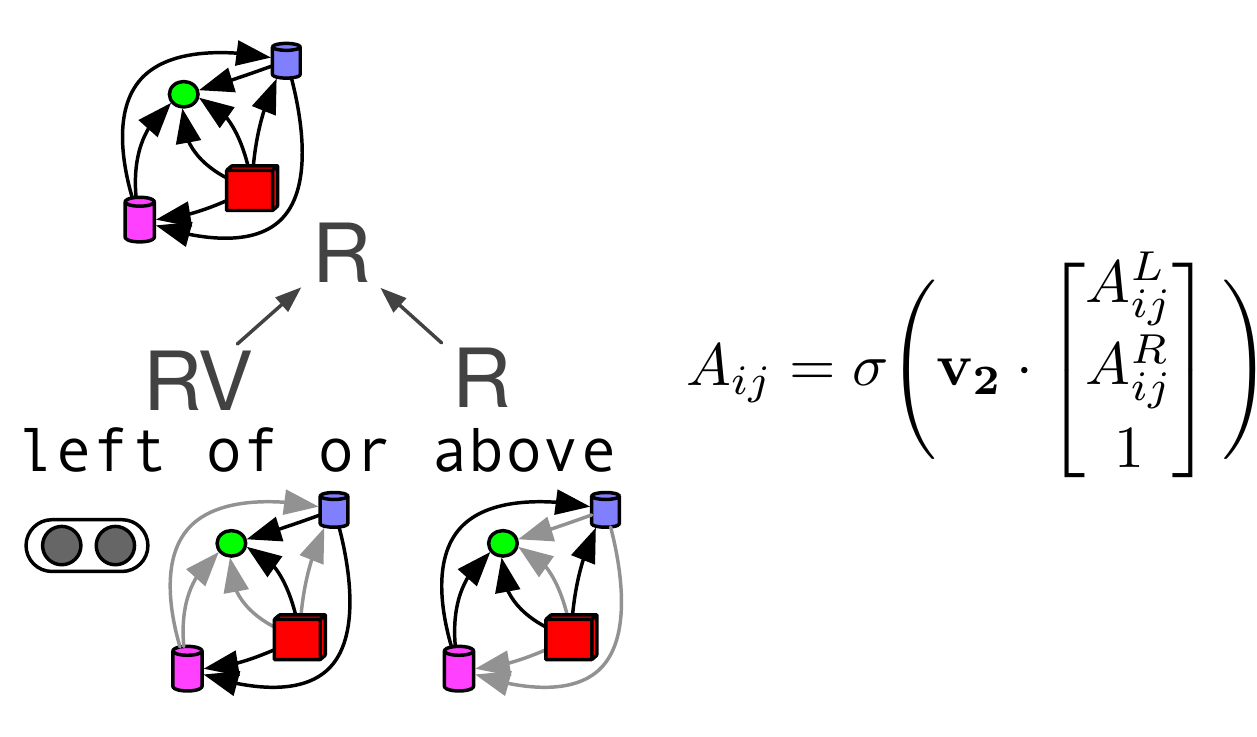}
    \caption{\label{fig:rwrr}{\bf \st{RV} + \st{R} $\rightarrow$ \st{R}}: This module composes a pair of soft set of relations to a produce an output soft set of relations. For example, the relations \emph{left} and \emph{above} are composed by the word \emph{or} to produce a set of relations such that entities $e_{i}$ and $e_{j}$ are related if either of the two relations exists between them. The functional form for this composition is similar to \st{EV}~+~\st{E}~$\rightarrow$~\st{E} and \st{TV}~+~\st{T}~$\rightarrow$~\st{T} modules. }
  \end{subfigure}
  
  \caption{\label{fig:modules} Composition Modules that compose two constituent span representations into the representation for the combined larger span, using the indicated equations. 
  }
\end{figure*}

Next, we describe how we compose phrase representations (from \S~\ref{ssec:semtype}) to represent larger phrases. 
We define a small set of composition modules, that take as input two constituents of text with their corresponding semantic representations (grounded representations and ungrounded vectors), and outputs the semantic type and corresponding representation of the larger constituent.
The composition modules are parameterized by the trainable word vectors.
These can be divided into several categories:

\paragraph{Composition modules resulting in fully grounded denotations:} Described in Figure~\ref{fig:modules}.

\paragraph{Composition with $\pmb{\phi}$-typed nodes:} Phrases with type $\pmb{\phi}$ are treated as being semantically transparent identity functions. Phrases of any other type can combined with these nodes, with no change to their type or representation.

\paragraph{Composition modules resulting in partially grounded denotations:} We define several modules that combine fully grounded phrases with ungrounded phrases, by deterministically taking the union of the representations, giving phrases with partially grounded representations (\S~\ref{ssec:semtype}). These modules are useful when words act as binary functions; here they combine with their first argument. For example, in Fig.~\ref{fig:overview}, \emph{or} and \emph{not cylindrical} combine to make a phrase containing both the vectors for \emph{or} and the entity set for \emph{not cylindrical}.

\section{Parsing Model}
\label{ssec:parsing}
Here, we describe how our model classifies question tokens into semantic type spans and computes their representations (\S~\ref{sssec:leaf}), and recursively uses the composition modules defined above to parse the question into a soft latent tree that provides the answer (\S~\ref{sssec:parse}). The model is trained end-to-end using only question-answer supervision (\S~\ref{sssec:train}).

\subsection{Lexical Representation Assignment}
\label{sssec:leaf}
Each token in the question sentence is assigned a distribution over the semantic types, and a grounded representation for each type. Tokens can only be assigned the \st{E}, \st{R}, \st{V}, and $\pmb{\phi}$ types.
For example, the token \emph{cylindrical} in the question in Fig.~\ref{fig:overview} is assigned a distribution over the $4$ semantic types (one shown) and for the \st{E} type, its representation is the set of {\it cylindrical} entities.

\paragraph{Semantic Type Distribution for Tokens:} To compute the semantic type distribution, our model represents each word $w$, and each semantic type $t$ using an embedding vector; $v_{w}, v_{t} \in \mathbb{R}^{d}$. The semantic type distribution is assigned with a softmax:
\begin{equation*}
    p(t|w_{i}) \propto \exp({v_{t}\cdot v_{w_{i}}})
    \label{eq:wordtypdist}
\end{equation*}

\paragraph{Grounding for Tokens:}
For each of the semantic type, we need to compute their representations:
\begin{enumerate}
    \item
    \st{E}-Type Representation: Each entity $e \in \E$, is represented using an embedding vector $v_{e} \in \mathbb{R}^{d}$ based on the concatenation of vectors for its properties. For each token $w$, we use its word vector to find the probability of each entity being part of the \st{E}-Type grounding:
    \begin{equation*}
        p^{w}_{e_{i}} = \sigma(v_{e_{i}} \cdot v_{w}) \; \; \forall \; e_{i} \in \E
    \end{equation*}
    For example, in Fig.~\ref{fig:overview}, the word \emph{red} will be grounded as all the red entities.

    \item
    \st{R}-Type Representation: Each relation $r \in \R$, is represented using an embedding vector $v_{r} \in \mathbb{R}^{d}$. For each token $w_{i}$, we compute a distribution over relations, and then use this to compute the {\it expected} adjacency matrix that forms the \st{R}-type representation for this token.
    \begin{align*}
        p(r|w_{i}) & \propto \exp({v_{r}\cdot v_{w_{i}}}) \\
        A^{w_{i}} & = \sum\limits_{r \in \R} p(r|w_{i})\cdot A_{r}
    \end{align*}
    e.g.\ the word \emph{left} in Fig.~\ref{fig:overview} is grounded as the subset of edges with label `left'.

    \item
    \st{V}-Type Representation: For each word $w \in \mathcal{V}$, we learn four vectors $v_{1} \in \mathbb{R}^{2}, v_{2} \in \mathbb{R}^{3}, v_{3} \in \mathbb{R}^{4}, v_{4} \in \mathbb{R}^{5}$, and use these as the representation for words with the \st{V}-Type.

    \item
    $\pmb{\phi}$-Type Representation: Semantically vacuous words that do not require a representation. 
\end{enumerate}

\subsection{Parsing Questions}
\label{sssec:parse}.
To learn the correct structure for applying composition modules, we use a simple parsing model. We build a parse-chart over the question encompassing all possible trees by applying all composition modules, similar to a standard CRF-based PCFG parser using the CKY algorithm. 
Each node in the parse-chart, for each span $w_{i..j}$ of the question, is represented as a distribution over different semantic types with their corresponding representations. 

\paragraph{Phrase Semantic Type Potential ($\phiijt{i}{j}{t}$):}
The model assigns a score, $\phiijt{i}{j}{t}$, to each $w_{i..j}$ span, for each semantic type \st{t}.
This score is computed from all possible ways of forming the span $w_{i..j}$ with type \st{t}.
For a particular composition of span $w_{i..k}$ of type $\mathbf{t_{1}}$ and $w_{k+1..j}$ of type $\mathbf{t_{2}}$, using the $\mathbf{t_{1}}+\mathbf{t_{2}}\rightarrow\mathbf{t}$ module, the composition score is:
\begin{align*}
\phiijkmod  =  \phiijt{i}{k}{t_{1}} \cdot \phiijt{k+1}{j}{t_{2}} 
    \cdot e^{\theta\cdot f^{t_{1} + t_{2} \rightarrow t}(i, j, k | q)}
\end{align*}
where $\theta$ is a trainable vector and $f^{t_{1} + t_{2} \rightarrow t}(i, j, k | q)$ is a simple feature function. Features consist of a conjunction of the composition module type and: the words before ($w_{i-1}$) and after ($w_{j+1}$) the span, the first ($w_{i}$) and last word ($w_{k}$) in the left constituent, and the first ($w_{k+1}$) and last ($w_{j}$) word in the right constituent.

\noindent
The final \st{t}-type potential of $w_{i..j}$ is computed by summing scores over all possible compositions:
\begin{equation*}
    \phiijt{i}{j}{t} = \sum\limits_{k = i}^{j-1} \sum\limits_{\substack{(t_{1} + t_{2} \rightarrow t) \\\in \text{Modules}}} \phiijkmod 
\end{equation*}
\paragraph{Combining Phrase Representations ($\llbracket w_{i..j} \rrbracket_{KG}^{t}$):}
To compute $w_{i..j}$'s \st{t}-type denotation, $\llbracket w_{i..j} \rrbracket_{KG}^{t}$, we compute an expected output representation from all possible compositions that result in type \st{t}.
\begin{align*}
\llbracket w_{i..j} \rrbracket_{KG}^{t} & = \frac{1}{\phiijt{i}{j}{t}} \sum\limits_{k = i}^{j-1} \llbracket w_{i..k..j} \rrbracket_{KG}^{t} \\
\llbracket w_{i..k..j} \rrbracket_{KG}^{t} & =  \sum\limits_{\substack{(t_{1} + t_{2} \rightarrow t) \\\in \text{Modules}}}\! \phiijkmod\! \cdot \!\llbracket w_{i..k..j} \rrbracket_{KG}^{t_{1} + t_{2} \rightarrow t}
\end{align*}
where $\llbracket w_{i..j} \rrbracket_{KG}^{t}$, is the \st{t}-type representation of the span $w_{i..j}$ and $\llbracket w_{i..k..j} \rrbracket_{KG}^{t_{1} + t_{2} \rightarrow t}$ is the representation resulting from the composition of $w_{i..k}$ with $w_{k+1..j}$ using the $\mathbf{t_{1}}+\mathbf{t_{2}}\rightarrow\mathbf{t}$ composition module.

\paragraph{Answer Grounding:}
By recursively computing the phrase semantic-type potentials and representations, we can infer the semantic type distribution of the complete question and the resulting grounding for different semantic types $t$, $\llbracket w_{1..|q|} \rrbracket_{KG}^{t}$.
\begin{equation}
\label{eq:qtd}
    p(t|q) \propto \psi(1, |q|, t)
\end{equation}
The answer-type (boolean or subset of entities) for the question is computed using:
\begin{equation}
    t^{*} = \argmax_{t \in {\st{T}, \st{E}}} \; p(t|q)
\end{equation}
The corresponding grounding is $\llbracket w_{1..|q|} \rrbracket_{KG}^{t^{*}}$, which answers the question.

\subsection{Training Objective}
\label{sssec:train}
Given a dataset $\mathcal{D}$ of (question, answer, knowledge-graph) tuples, $\{\text{q}^{i}, \text{a}^{i}, \text{KG}^{i}\}_{i=1}^{i=|\mathcal{D}|}$, we train our model to maximize the log-likelihood of the correct answers.
We maximize the following objective:
\begin{align}
\mathcal{L} &= \sum_i \log p(\text{a}^{i} | \text{q}^{i}, \text{KG}^{i})
\end{align}
Further details regarding the training objective are given in Appendix A.

\begin{table*}[tb]

\centering
\small
\begin{tabular}{l c c c c} 
\toprule
\bf Model & \bf Boolean Questions & \bf Entity Set Questions & \bf Relation Questions & \bf Overall \\
\midrule
\textsc{LSTM (No KG)}               & 50.7 & 14.4 & 17.5  & 27.2 \\
\textsc{LSTM }                      & 88.5 & 99.9 & 15.7  & 84.9 \\
\textsc{Bi-LSTM }                   & 85.3 & 99.6 & 14.9  & 83.6 \\
\textsc{Tree-LSTM}                  & 82.2 & 97.0 & 15.7  & 81.2 \\
\textsc{Tree-LSTM (Unsup.)}         & 85.4 & 99.4 & 16.1  & 83.6 \\
\textsc{Relation Network}           & 85.6 & 89.7 & 97.6  & 89.4 \\
\addlinespace[1mm]
Our Model (Pre-parsed)              & 94.8 & 93.4   & 70.5  & 90.8 \\
Our Model                           & 99.9 & 100    & 100   & 99.9 \\
\bottomrule
\end{tabular}
\caption{\label{tab:qa}\textbf{Results for Short Questions ({\sc CLEVRGEN})}: Performance of our model compared to baseline models on the Short Questions test set. The \textsc{LSTM (No KG)} has accuracy close to chance, showing that the questions lack trivial biases. 
Our model almost perfectly solves all questions showing its ability to learn challenging semantic operators, and parse questions only using weak end-to-end supervision.}
\end{table*}

\section{Dataset}
\label{sec:dataset}

We experiment with two datasets, 1) Questions generated based on the {\sc CLEVR} \citep{Johnson_2017_CVPR} dataset, and 2) Referring Expression Generation (GenX) dataset \citep{FitzGerald2013LearningDO}, both of which feature complex compositional queries.

\paragraph{\textsc{CLEVRGEN}:} 
We generate a dataset of question and answers based on the {\sc CLEVR} dataset \citep{Johnson_2017_CVPR}, which contains knowledge graphs containing attribute information of objects and relations between them. 

We generate a new set of questions as existing questions contain some biases that can be exploited by models.\footnote{
\newcite{Johnson_2017_CVPR} found that many spatial relation questions can be answered only using absolute spatial information, and many long questions can be answered correctly without performing all steps of reasoning. We employ some simple tests to remove trivial biases from our dataset.}
We generate 75K questions for training and 37.5K for validation.
Our questions test various challenging semantic operators.
These include conjunctions (e.g. \emph{Is anything red and large?}), negations (e.g. \emph{What is not spherical?}), counts (e.g. \emph{Are five spheres green?}), quantifiers (e.g. \emph{Is every red thing cylindrical?}), and relations (e.g. \emph{What is left of and above a cube?}). We create two test sets:

\begin{enumerate}
\item
\textbf{Short Questions:}
Drawn from the same distribution as the training data (37.5K).

\item
\textbf{Complex Questions:}
Longer questions than the training data (22.5K). This test set contains the same words and constructions, but chained into longer questions.
For example, it contains questions such as \emph{What is a cube that is right of a metallic thing that is beneath a blue sphere?} and \emph{Are two red cylinders that are above a sphere metallic?}
Solving these questions require more multi-step reasoning.
\end{enumerate}

\paragraph{\textsc{Referring Expressions (GenX)}}\citep{FitzGerald2013LearningDO}: This dataset contains human-generated queries, which identify a subset of objects from a larger set (e.g. \emph{all of the red items except for the rectangle}).
It tests the ability of models to precisely understand human-generated language, which contains a far greater diversity of syntactic and semantic structures.
This dataset does not contain relations between entities, and instead only focuses on entity-set operations.
The dataset contains 3920 questions for training, 600 for development and 940 for testing. 
Our modules and parsing model were designed independently of this dataset, and we re-use hyperparameters from \textsc{CLEVRGEN}.

\section{Experiments}
\label{sec:exp}
Our experiments investigate the ability of our model to understand complex synthetic and natural language queries, learn interpretable structure, and generalize compositionally.
We also isolate the effect of learning the syntactic structure and representing sub-phrases using explicit denotations.

\subsection{Experimentation Setting}
We describe training details, and the baselines.
\paragraph{Training Details:}
Training the model is challenging since  it needs to learn both good syntactic structures and the complex semantics of neural modules---so we use Curriculum Learning \citep{bengio2009curriculum} to pre-train the model on an easier subset of questions. Appendix B contains the details of curriculum learning and other training details.

\paragraph{Baseline Models:}
We compare to the following baselines.
\textbf{(a)} Models that assume linear structure of language, and encode the question using linear RNNs---\textsc{LSTM (No KG)}, \textsc{LSTM}, \textsc{Bi-LSTM}, and a \textsc{Relation-Network}~\citep{SantoroRBMPBL17} augmented model.
\footnote{We use this baseline only for {\sc CLEVRGEN} since {\sc GenX} does not contain relations.}
\textbf{(b)} Models that assume tree-like structure of language. We compare two variants of Tree-structured LSTMs~\cite{zhu2015long, Tai2015ImprovedSR}---\textsc{Tree-LSTM}, that operates on pre-parsed questions, and \textsc{Tree-LSTM(Unsup.)}, an unsupervised Tree-LSTM model~\cite{Maillard2017} that learns to jointly parse and represent the sentence.
For {\sc GenX}, we also use an end-to-end semantic parsing model from \citet{Pasupat2015CompositionalSP}. 
Finally, to isolate the contribution of the proposed denotational-semantics model, we train our model on pre-parsed questions. 
Note that, all LSTM based models only have access to the entities of the KG but not the relationship information between them. 
See Appendix C for details.

\begin{table}[tb]
\centering
\small
\begin{tabular}{l@{\hskip 0.0cm} c@{\hskip 0.1cm} c@{\hskip 0.1cm} c}
\toprule
{\bf Model}                & {\bf \begin{tabular}{@{}c@{}}Non-relation \\ Questions\end{tabular}}        & {\bf \begin{tabular}{@{}c@{}}Relation \\ Questions\end{tabular}}        & {\bf Overall } \\
\midrule
\textsc{LSTM (No KG)}               & 46.0               & 39.6                & 41.4  \\
\textsc{LSTM }                      & 62.2               & 49.2                & 52.2  \\
\textsc{Bi-LSTM}                    & 55.3               & 47.5                & 49.2  \\
\textsc{Tree-LSTM}                  & 53.5               & 46.1                & 47.8  \\
\textsc{Tree-LSTM (Unsup.)}         & 64.5               & 42.6                & 53.6  \\
\textsc{Relation Network}           & 51.1               & 38.9                & 41.5  \\
\addlinespace[1mm]
Our Model (Pre-parsed)              & 94.7               & 74.2                & 78.8  \\
Our Model                           & 81.8               & 85.4                & 84.6  \\
\bottomrule
\end{tabular}
\caption{\label{tab:longerqa}\textbf{Results for Complex Questions ({\sc CLEVRGEN})}: All baseline models fail to generalize well to questions requiring longer chains of reasoning than those seen during training. Our model substantially outperforms the baselines, showing its ability to perform complex multi-hop reasoning, and generalize from its training data. Analysis suggests that most errors from our model are due to assigning incorrect structures, rather than mistakes by the composition modules. }
\end{table}

\subsection{Experiments}

\paragraph{Short Questions Performance:} Table~\ref{tab:qa} shows that our model perfectly answers all test questions, demonstrating that it can learn challenging semantic operators and induce parse trees from end task supervision.
Performance drops when using external parser, showing that our model learns an effective syntactic model for this domain.
The \textsc{Relation Network} also achieves good performance, particularly on questions involving relations.
LSTM baselines work well on questions not involving relations.\footnote{Relation questions are out of scope for these models.}

\paragraph{Complex Questions Performance:}
Table~\ref{tab:longerqa} shows results on complex questions, which are constructed by combining components of shorter questions.
These require complex multi-hop reasoning, and the ability to generalize robustly to new types of questions.
We use the same models as in Table~\ref{tab:qa}, which were trained on short questions.
All baselines achieve close to random performance, despite high accuracy for shorter questions. This shows the challenges in generalizing RNN encoders
beyond their training data.
In contrast, the strong inductive bias from our model structure allows it to generalize well to  complex questions.
Our model outperforms \textsc{Tree-LSTM (Unsup.)} and the version of our model that uses pre-parsed questions, showing the effectiveness of explicit denotations and learning the syntax, respectively.

\begin{table}[tb]
\centering
\small
\begin{tabular}{l c}
\toprule
{\bf Model}                & {\bf Accuracy} \\
\midrule
\textsc{LSTM (No KG)}          & 0.0 \\
\textsc{LSTM}                  & 64.9 \\
\textsc{Bi-LSTM}               & 64.6 \\
\textsc{Tree-LSTM}             & 43.5 \\
\textsc{Tree-LSTM (Unsup.)}    & 67.7 \\
\addlinespace[1mm]
\textsc{Sempre}                & 48.1 \\
\addlinespace[1mm]
Our Model (Pre-parsed)         & 67.1 \\
Our Model                      & 73.7 \\
\bottomrule
\end{tabular}
\caption{\label{tab:regqa}\textbf{Results for Human Queries ({\sc GenX})} Our model outperforms \textsc{LSTM} and semantic parsing models on complex human-generated queries, showing it is robust to work on natural language. Better performance than \textsc{Tree-LSTM (Unsup.)} shows the efficacy in representing sub-phrases using explicit denotations. Our model also performs better without an external parser, showing the advantages of latent syntax.}

\end{table}

\paragraph{Performance on Human-generated Language:}
Table \ref{tab:regqa} shows the performance of our model on complex human-generated queries in {\sc GenX}.
Our approach outperforms strong LSTM and semantic parsing baselines, despite the semantic parser's use of hard-coded operators.
These results suggest that our method represents an attractive middle ground between minimally structured and highly structured approaches to interpretation.
Our model learns to interpret operators such as \emph{except} that were not considered during development.
This shows that our model can learn to parse human language, which contains greater lexical and structural diversity than synthetic questions. Trees induced by the model are linguistically plausible (see Appendix D).

\paragraph{Error Analysis:} We find that most model errors are due to incorrect assignments of structure, rather than semantic errors from the modules. For example, in the question \emph{Are four red spheres beneath a metallic thing small?}, our model's parse composes \emph{metallic thing small} into a constituent instead of composing \emph{red spheres beneath a metallic thing} into a single node. 
Future work should explore more sophisticated parsing models.

\paragraph{Discussion:} While our model shows promising results, there is significant potential for future work. Performing exact inference over large KGs is likely to be intractable, so approximations such as KNN search, beam search, feature hashing or parallelization may be necessary.  To model the large number of entities in KGs such as Freebase, techniques proposed by recent work~\citep{VergaRowless, GuptaSiRo17} that explore representing entities as composition of its properties, such as, types, description etc. could be used..
The modules in this work were designed in a way to provide good inductive bias for the kind of composition we expected them to model. For example,  \st{EV}~+~\st{E}~$\rightarrow$~\st{E} is modeled as a linear composition function making it easy to represent words such as \emph{and} and \emph{or}.
These modules can be exchanged with any other function with the same `type signature', with different trade-offs---for example, more general feed-forward networks with greater representation capacity would be needed to represent a linguistic expression equivalent to \emph{xor}.
Similarly, more module types would be required to handle certain constructions---for example, a multiword relation such as \emph{much larger than} needs a \st{V}~+~\st{V}~$\rightarrow$~\st{V} module. 
This is an exciting direction for future research.

\section{Related Work}
\label{sec:relwork}
Many approaches have been proposed to perform question-answering against structured knowledge sources. \emph{Semantic parsing} models have learned structures over pre-defined discrete operators, to produce logical forms that can be executed to answer the question. Early work trained using gold-standard logical forms \citep{zettlemoyer2005learning, Kwiatkowski2010InducingPC}, whereas later efforts have only used answers to questions \citep{Clarke2010DrivingSP, liang2011learning, Krishnamurthy2013JointlyLT}. 
A key difference is that our model must learn semantic operators from data, which may be necessary to model the fuzzy meanings of function words like \emph{many} or \emph{few}.

Another similar line of work is neural program induction models, such as Neural Programmer~\citep{Neelakantan2016LearningAN} and Neural Symbolic Machine~\citep{Liang2017NeuralSM}. These models learn to produce programs composed of predefined operators using weak supervision to answer questions against semi-structured tables.

Neural module networks have been proposed for learning semantic operators \citep{andreas2016neural} for question answering. This model assumes that the structure of the semantic parse is given, and must only learn a set of operators. Dynamic Neural Module Networks (D-NMN) extend this approach by selecting from a small set of candidate module structures \citep{Andreas2016LearningTC}. We instead learn a model over all possible structures. 

Our work is most similar to N2NMN \citep{Hu2017LearningTR} model, which learns both semantic operators and the layout in which to compose them. However, optimizing the layouts requires reinforcement learning, which is challenging due to the high variance of policy gradients, whereas our chart-based approach is end-to-end differentiable. 

\section{Conclusion}
\label{sec:conclusion}
We have introduced a model for answering questions requiring compositional reasoning that combines ideas from compositional semantics with end-to-end learning of composition operators and structure. We demonstrated that the model is able to learn a number of complex composition operators from end task supervision, and showed that the linguistically motivated inductive bias imposed by the structure of the model allows it to generalize well beyond its training data.
Future work should explore scaling the model to other question answering tasks, using more general composition modules, and introducing additional module types.

\section*{Acknowledgement}
We would like to thank Shyam Upadhyay and the anonymous reviewers for their helpful suggestions.

\bibliographystyle{acl_natbib_nourl}
\bibliography{references}

\appendix

\section{Training Objective}
Given a dataset $\mathcal{D}$ of (question, answer, knowledge-graph) tuples, $\{\text{q}^{i}, \text{a}^{i}, \text{KG}^{i}\}_{i=1}^{i=|\mathcal{D}|}$, we train our model to maximize the log-likelihood of the correct answers.
Answers are either booleans ($a \in \{0, 1\}$), or specific subsets of entities ($a = \{e_{j}\}$) from the KG.
We denote the semantic type of the answer as $a_{t}$. 
The model's answer is found by taking the complete question representation, containing a distribution over types and the representation for each type.
We maximize the following objective:
\begin{align}
\mathcal{L} &= \sum_i \log p(\text{a}^{i} | \text{q}^{i}, \text{KG}^{i}) \\
            &= \sum_{i} \mathcal{L}^{i}_{b} + \mathcal{L}^{i}_{e}
\end{align}
where $\mathcal{L}^{i}_{b}$ and $\mathcal{L}^{i}_{e}$ are respectively the objective functions for questions with boolean answers and entity set answers.
\begin{equation}
\mathcal{L}^{i}_{b}  = \mathbbm{1}_{a^{i}_{t}=\st{T}} \Big[\log (p_{true})^{a^{i}} (1-p_{true})^{(1-a^{i})} \Big] 
\end{equation}
\begin{equation}
\mathcal{L}^{i}_{e} = \frac{\mathbbm{1}_{a^{i}_{t}=\st{E}}}{|\E^{i}|} 
                    \Big[ \log \prod\limits_{e^{i}_{j} \in a^{i}} p_{e^{i}_{j}} \prod\limits_{e^{i}_{j} \notin a^{i}} (1 - p_{e^{i}_{j}})\Big]
\end{equation}
We also add $L_{2}$-regularization for the scalar parsing features introduced in \S~\ref{sssec:parse}.

\section{Training Details}
\paragraph{Representing Entities:} Each entity in {\sc CLEVRGEN} and {\sc GenX} datasets consists of $4$ attributes. For each attribute-value, we learn an embedding vector and concatenate these vectors to form the representation for the entity.

\paragraph{Training details:}
For curriculum learning, for the {\sc CLEVRGEN} dataset we use a 2-step schedule where we first train our model on simple attribute match (\emph{What is a red sphere?}), attribute existence (\emph{Is anything blue?}) and boolean composition (\emph{Is anything green and is anything purple?}) questions and in the second step on all questions. For {\sc GenX} we use a 5-step, question-length based schedule, where we first train on shorter questions and eventually on all questions.

We tune hyper-parameters using validation accuracy on the {\sc CLEVRGEN} dataset, and use the same hyper-parameters for both datasets. We train using SGD with a learning rate of $0.5$, a mini-batch size of 4, and regularization constant of $0.3$.
When assigning the semantic type distribution to the words at the leaves, we add a small positive bias of $+1$ for $\pmb{\phi}$-type and a small negative bias of $-1$ for the \st{E}-type score before the softmax.
Our trainable parameters are: question word embeddings ($64$-dimensional), relation embeddings ($64$-dimensional), entity attribute-value embeddings ($16$-dimensional), four vectors per word for \st{V}-type representations, parameter vector $\theta$ for the parsing model that contains six scalar feature scores per module per word, and the global parameter vector for the \st{E}+\st{E}$\rightarrow$\st{E} module.

\section{Baseline Models}

\subsection{LSTM (No KG)}
We use a LSTM network to encode the question as a vector $q$. We also define three other parameter vectors, $t$, $e$ and $b$ that are used to predict the answer-type $P(a=\st{T}) = \sigma(q\cdot t)$, entity attention value $p_{e_{i}} = \sigma(q\cdot e)$, and the probability of the answer being True $p_{true} = \sigma(q\cdot b)$.

\subsection{LSTM}
Similar to \textsc{LSTM (No Relation)}, the question is encoded using a LSTM network as vector $q$. Similar to our model, we learn entity attribute-value embeddings and represent each entity as the concatenation of the $4$ attribute-value embeddings, $v_{e_{i}}$. Similar to \textsc{LSTM (No Relation)}, we also define the $t$ parameter vector to predict the answer-type.
The entity-attention values are predicted as $p_{e_{i}} = \sigma(v_{e_{i}}\cdot q)$.
To predict the probability of the boolean-type answer being true, we first add the entity representations to form $b = \sum\limits_{e_{i}}v_{e_{i}}$, then make the prediction as $p_{true} = \sigma(q\cdot b)$.

\subsection{Tree-LSTM}
Training the Tree-LSTM model requires pre-parsed sentences for which we use a binary constituency tree generating PCFG parser~\cite{klein2003accurate}. We input the pre-parsed question to the Tree-LSTM to get the question embedding $q$. The rest of the model is same the  \textsc{LSTM} model above.

\subsection{Relation Network Augmented Model}
The original formulation of the relation network module is as follows:
\begin{equation}
    RN(q, KG) = f_{\phi}\Bigg( \sum\limits_{i, j} g_{\theta} (e_{i}, e_{j}, q) \Bigg) 
\end{equation}
where $e_{i}$, $e_{j}$ are the representations of the entities and $q$ is the question representation from an LSTM network.
The output of the Relation Network module is a scalar score value for the elements in the answer vocabulary. 
Since our dataset contains entity-set valued answers, we modified the module in the following manner.

We concatenate the entity-pair representations with the representations of the pair of relations between them\footnote{In the {\sc CLEVR} dataset, between any pair of entities, only 2 directed relations, \emph{left} or \emph{right}, and \emph{above} or \emph{beneath} are present.}.
We use the RN-module to produce an output representation for each entity as: 
\begin{equation}
    RN_{e_{i}} = f_{\phi}\Bigg( \sum\limits_{j} g_{\theta} (e_{i}, e_{j}, r^{1}_{ij}, r^{2}_{ij}, q) \Bigg)
\end{equation}
Similar to the \textsc{LSTM} baselines, we define a parameter vector $t$ to predict the answer-type, and compute the vector $b$ to compute the probability of the boolean type answer being true.

To predict the entity-attention values, we use a separate attribute-embedding matrix to first generate the output representation for each entity, $e^{out}_{i}$, then predict the output attention values as follows:
\begin{equation}
    p_{e_{i}} = \sigma\bigg(RN_{e_{i}}\cdot e^{out}_{i}\bigg)
\end{equation}

We tried other architectures as well, but this modification provided the best performance on the validation set. We also tuned the hyper-parameters and found the setting from \citet{SantoroRBMPBL17} to work the best based on validation accuracy.
We used a different 2-step curriculum to train the {\sc Relation Network} module, in which we replace the Boolean questions with the relation questions in the first-schedule and jointly train on all questions in the subsequent schedule.

\subsection{\textbf{\textsc{Sempre}}}
The semantic parsing model from \citep{Pasupat2015CompositionalSP} answers natural language queries for semi-structured tables. The answer is a denotation as a list of cells in the table. To use the \textsc{Sempre} framework, we convert the KGs in the \textsc{GenX} to tables as follows:
\begin{enumerate}
\item Each table has the first row (header) as: $|\;\; ObjId \;\;|\;\; P1\;\; |\;\; P2 \;\;|\;\; P3\;\; |\;\; P4\;\;|$

\item Each row contains an object id, and the 4 property-attribute values in cells.

\item The answer denotation, i.e. the objects selected by the human annotators is now represented as a list of object ids.
\end{enumerate}

After converting the the KGs to tables, \textsc{Sempre} framework can be trivially used to train and test on the \textsc{GenX} dataset. We tune the number of epochs to train for based on the validation accuracy and find $8$ epochs over the training data to work the best. We use the default setting of the other hyper-parameters.

\section{Example Output Parses}
\label{appendix:parses}
\begin{figure*}[ht]
    \centering
    \begin{subfigure}[t]{\linewidth}
        \centering
        \includegraphics[width=\linewidth]{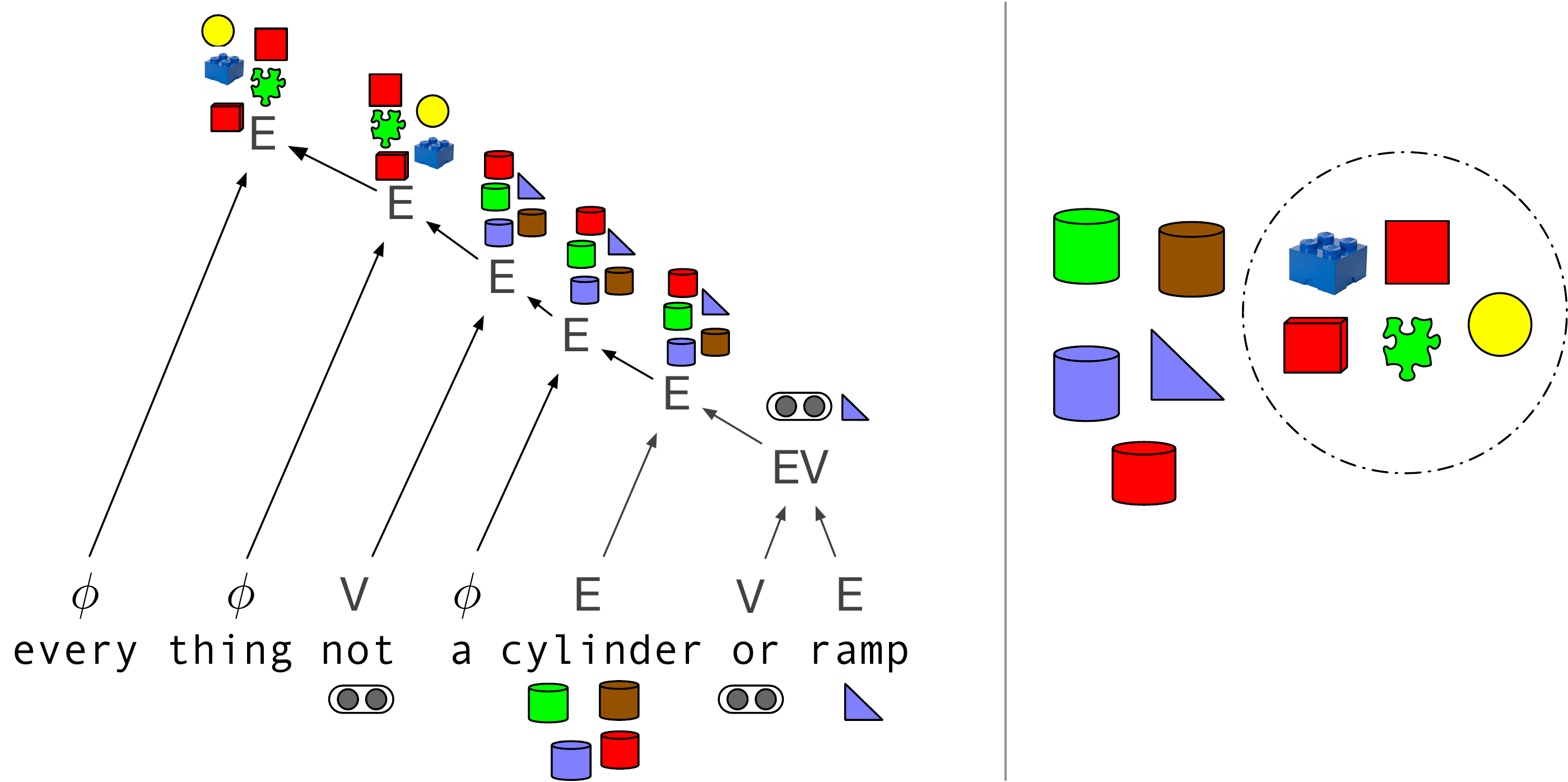}
        \caption{An example output from our learned model showing that our model learns to correctly parse the questions sentence, and model the relevant semantic operator; \emph{or} as a set union operation, to generate the correct answer denotation. It also learns to cope with lexical variability in human language; \emph{triangle} being referred to as \emph{ramp}.}
    \end{subfigure}%
    
    \medskip

    \begin{subfigure}[t]{\linewidth}
        \centering
        \includegraphics[width=\linewidth]{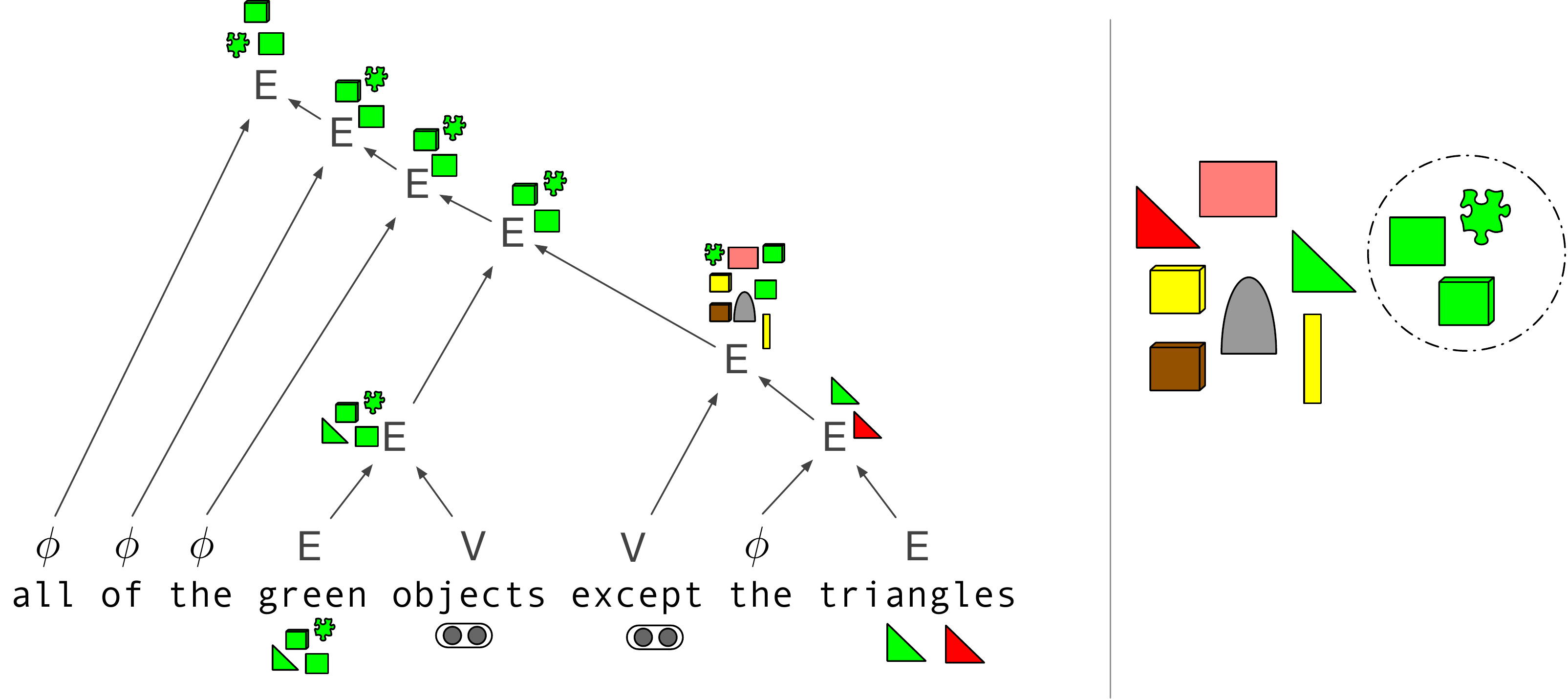}
        \caption{An example output from our learned model that shows that our model can learn to correctly parse human-generated language into relatively complex structures and model semantic operators, such as \emph{except}, that were not encountered during model development.}
    \end{subfigure}%
    
    \caption{\textbf{Example output structures from our learned model}: Examples of queries from the \textsc{GenX} dataset and the corresponding highest scoring tree structures from our learned model. The examples shows that our model is able to correctly parse human-generated language and jointly learn to model semantic operators, such as set unions, negations etc.}
    
    \label{fig:example_parses}
\end{figure*}

In Figure~\ref{fig:example_parses}, we show example queries and their highest scoring output structure from our learned model for \textsc{GenX} dataset.

\end{document}